\documentclass[sigconf, natbib=true, anonymous=false]{acmart}

\usepackage{graphicx}
\usepackage{booktabs}
\usepackage{multirow}
\usepackage{amsmath}
\usepackage{enumitem}
\usepackage{xcolor}
\usepackage{hyperref}
\usepackage{listings}
\usepackage{courier}
\usepackage{marvosym}

\lstdefinelanguage{Cypher}{
  morekeywords={MATCH,WHERE,RETURN,AND,OR,NOT,CONTAINS},
  sensitive=true,
  morecomment=[l]{//},
  morestring=[b]",
}

\renewcommand\footnotetextcopyrightpermission[1]{}
\title{Querying Climate Knowledge: Semantic Retrieval for Scientific Discovery}

\author{Mustapha Adamu\textsuperscript{1}, Qi Zhang\textsuperscript{1}, Huitong Pan\textsuperscript{1}, Longin Jan Latecki\textsuperscript{1}, Eduard C. Dragut\textsuperscript{1} \textsuperscript{\Letter}}

\affiliation{\textsuperscript{1}Temple University, Philadelphia, PA, USA\country{USA}}

\email{{mustapha.adamu, qi.zhang, huitong.pan, latecki, edragut}@temple.edu}

\vspace{-12pt}
\begin{document}
\begin{abstract}
The growing complexity and volume of climate science literature make it increasingly difficult for researchers to find relevant information across models, datasets, regions, and variables. This paper introduces a domain-specific Knowledge Graph (KG) built from climate publications and broader scientific texts, aimed at improving how climate knowledge is accessed and used. Unlike keyword-based search, our KG supports structured, semantic queries that help researchers discover precise connections—such as which models have been validated in specific regions or which datasets are commonly used with certain teleconnection patterns. We demonstrate how the KG answers such questions using Cypher queries, and outline its integration with large language models in RAG systems to improve transparency and reliability in climate-related question answering. This work moves beyond KG construction to show its real-world value for climate researchers, model developers, and others who rely on accurate, contextual scientific information.
\end{abstract}
\maketitle

\section*{Introduction}

Climate science continues to produce vast amounts of data and research, from global climate model simulations to regional impact assessments and peer-reviewed literature \citep{pan2024climatener, zhang2024scier, wu2022linkclimate}. 
For both researchers and policymakers, navigating this expanding knowledge base has become increasingly challenging. Traditional keyword-based search tools often prove too broad or imprecise, returning results that lack the contextual nuance necessary to compare studies meaningfully. For example, a search for ``\textit{sea level rise in coastal cities}'' on Google might yield hundreds of papers from various sources, yet offer no clear way to differentiate among emission scenarios, model generations, or observational methods used \citep{reinanda2020knowledge}.

The difficulty lies not only in the volume of information but also in its complexity. 
Climate science is fundamentally interdisciplinary, integrating atmospheric physics, oceanography, ecology, and the social sciences. Each domain brings its own specialized terminology, methodologies, and assumptions—making it difficult to trace connections or synthesize insights across studies \citep{defila2011defining, metoffice2022interdisciplinary, DynClean}. Simultaneously, the demand for transparency and reproducibility is growing: researchers must increasingly understand the provenance of datasets, models, and assumptions behind published findings \citep{schmidt2018reproducibility, rupprecht2020ursprung}. 
Tracking model genealogy is a key challenge in building an accurate climate knowledge graph. Many CMIP models share code or components, making it hard to determine which models are truly independent. This creates hidden dependencies in the ensemble and complicates how we represent model lineage and similarity in a structured way \citep{kruger2006nexrad,Kuma2023JAMES}.

To address these challenges, we construct a domain-specific climate knowledge graph, ClimatePub4KG, designed to support more structured, meaningful exploration of scientific literature about climate models and data. ClimatePub4KG captures relationships among key scientific entities—such as climate models, variables, locations, datasets, and teleconnection patterns—enabling users to pose queries that reflect real-world scientific reasoning. For example, one might ask: \textit{Which CMIP6 models project more than 2°C of warming by 2050 under SSP5-8.5, and have been validated using Arctic sea ice observations?}

We build on our prior efforts to showcase how ClimatePub4KG supports real-world scientific exploration \cite{dragut2024climatepub4kg}.
By integrating natural language queries with graph-based retrieval, we show how researchers can access precise and context-rich information more effectively.
Moreover, knowledge graphs like ClimatePub4KG can serve as a powerful retrieval backend for Retrieval-Augmented Generation (RAG) systems powered by large language models \cite{Xu2024, graphRAG}. 
While LLMs excel at language understanding and generation, they often struggle with factual consistency and reasoning over complex scientific information. By grounding LLMs in a domain-specific KG, we enable more accurate and verifiable responses, especially for multi-hop scientific questions that require integrating information across models, scenarios, and datasets.
For instance, a RAG system augmented with ClimatePub4KG can answer questions like ``\textit{Which CMIP6 models share same components with of NorESM2-LM, and how does that influence their ensemble projections over the Tropics?}'' with structured evidence—enhancing both the relevance and trustworthiness of the output. 
This integration of symbolic and neural methods represents a promising direction for scientific discovery and climate policy support. 
We present our use cases to engage the IR community and invite feedback on how to improve ClimatePub4KG and build novel applications with it.

\section*{Background and Related Work}


Knowledge graphs (KGs) have recently gained attention as a way to organize scientific knowledge. Projects like Semantic Scholar Semantic Scholar \cite{lo2019s2orc}, OpenAlex \cite{prieme2022openalex}, and AMiner \cite{tang2008arnetminer}
have shown how large-scale graphs help structure metadata and citation networks. But general-purpose KGs still struggle to support domain specific scientific queries—like understanding how specific models, variables, and outcomes relate—because they lack domain-specific structure.

Our earlier work addressed this gap by building ClimatePub4KG that captures models, datasets, weather events, teleconnection patterns, and the relationships between them. The ClimateIE framework \cite{pan2024climatener,pan2025climateie}  extracted structured data from climate publications using a domain taxonomy and expert-reviewed annotations. SciER \citep{zhang2024scier}  complemented this with entity and relation extraction across broader scientific text. Together, these efforts built the foundation for the structured ClimatePub4KG tailored to climate research.

Other projects—like Spider \citep{Yu2018Spider}, and work on question answering over KGs \citep{Hoffner2017KGQA, Cui2020KGQA} —show how natural language interfaces can be used to query structured data. Our work builds on that idea but focuses specifically on climate science, where terminology and questions are often more complex. This paper is not about how we built the ClimatePub4KG but rather about how it can now be used for domain-specific search and climate information retrieval.

\section*{Retrieval Augmented Generation}

The diverse and technical nature of climate science presents real challenges when using large language models. These models are trained on general datasets and often miss the depth and specificity needed for climate-related questions. As a result, they tend to produce unclear or even misleading answers when asked about models, experiments, or region specific impacts. To address this, we’ve developed a solution based on Retrieval Augmented Generation (RAG). Our approach grounds the language model with structured knowledge from trusted sources like GCMD and ESGF. By using a climate specific taxonomy and linking it to actual content from our knowledge graph, the system can return more accurate, focused, and transparent answers \cite{pan2025taxonomykg}. This not only improves reliability but also makes advanced tools like language models more useful to the climate research community. 

\section*{Use Case: Querying the Climate Knowledge Graph}
Here were present example queries to indicate how ClimatePub4KG will be beneficial to the climate science community. We will present 3 queries in Natural language, the relevant personas for the queries, its translated version into a structure language, like Cipher \citep{text2cypher} , along with brief explanations of the importance of the query. 

A climate researcher who for example needs information about a particular climate model to use for research might have to read several research papers sometimes without success to get details about climate model. Researches who also want to do ensemble research will need to know if there are dependencies between their models as that can lead to biased results. Non climate scientist such as civil engineers will still need climate information as part of their work and without climate science background finding the needed information could be very hectic. To this end, creating a ClimatePub4KG which includes various information and can be explored with simple queries could save climate scientist as well as relevant industry players enormous amount of time in the climate data discovery as well as delivery process. To this end our ClimatePub4KG seeks to provide avenue for solve most of the problems enumerated above.


A fundamental challenge in climate science is to identify and synthesize research that discusses specific phenomena within particular geographical and contextual constraints. For instance, a researcher investigating anomalous temperature regimes might pose the following question:

\textbf{Persona 1: Researcher}

Interest: To understand how temperature anomaly affect certain areas of the of the world. Below will be a natural language query that can be posed to our graph. 

\textbf{Natural Language Query 1:} ``Which papers mention anomalous temperature regimes such as cold air outbreaks (CAOs) or warm waves (WWs) in relation to North America, specifically in the sentences where these terms appear?''

The results of this query will produce a graph that shows weather events such as CAO, ENSO etc and link to places such as North America with supporting Literature to back this support. Enabling the person to find this information by running simple query instead of combining through large volume of data.

\begin{lstlisting}[language=Cypher, caption=Cypher Query 1]
MATCH (we:Weather_Event)-[:TargetsLocation]->(l:Location {Name:"NORTH_AMERICA"})
MATCH (p:Paper)-[m:Mention]->(we)
WHERE (m.Mention_Sentence CONTAINS 'WW' OR m.Mention_Sentence CONTAINS 'CAOs')
RETURN p.title AS PaperTitle, l.Name AS Location, 
       we.Name AS WeatherEvent, m.Mention_Sentence AS Context
\end{lstlisting}

This query navigates the graph by first identifying \texttt{Weather\_Event} nodes (aliased as \texttt{we}) that have a \texttt{TargetsLocation} relationship with a \texttt{Location} node named ``NORTH\_AMERICA'' (\texttt{l}). It then finds \texttt{Paper} nodes (\texttt{p}) that have a \texttt{Mention} relationship to these weather events. The \texttt{WHERE} clause filters these mentions to only those sentences containing the specific acronyms \texttt{'WW'} or \texttt{'CAOs'}. The \texttt{RETURN} clause then provides the paper title, location name, weather event name, and the contextual sentence. Such a query allows for highly targeted literature discovery, sifting through potentially thousands of papers to pinpoint those discussing specific phenomena in a relevant geographical context with high precision.

\textbf{Persona 2: Climate Model Devloper/Researcher}

Interest: Understanding how climate model developed are used and which papers cited them

Climate Model Intercomparison Projects (CMIPs) are central to modern climate science, providing standardized experimental protocols and model outputs. Understanding which models are used in conjunction with specific climate phenomena and regions is crucial for model evaluation and synthesis. Consider the query:

\textbf{Natural Language Query 3:} ``Which papers mention CMIP5 models and the North Atlantic Oscillation (NAO) in the context of the Southeast United States?''

This query combines information about a specific generation of climate models (CMIP5), a well-known teleconnection pattern (NAO), and a particular geographical region (Southeast United States). The Cypher translation is:


\begin{lstlisting}[language=Cypher, caption=Cypher Query 2]
MATCH (p:Paper)-[:Mention]->(mod:Model|Project)
WHERE mod.Name CONTAINS 'CMIP5'
MATCH (p)-[:Mention]->(tel:Teleconnection {Name:"NORTH_ATLANTIC_OSCILLATION"})
MATCH (p)-[:Mention]->(loc:Location)
WHERE loc.Name CONTAINS 'Southeast' AND 
      (loc.wikidata_description CONTAINS 'United States' OR 
       loc.Name CONTAINS 'US')
RETURN p.title AS PaperTitle, mod.Name AS ModelProject, 
       tel.Name AS Teleconnection, loc.Name AS Region
\end{lstlisting}

This query identifies \texttt{Paper} nodes (\texttt{p}) that mention \texttt{Model} or \texttt{Project} nodes (\texttt{mod}) containing \texttt{'CMIP5'} in their name. It further requires these papers to also mention the \texttt{Teleconnection} node representing the North Atlantic Oscillation (\texttt{tel}). Finally, it links these papers to \texttt{Location} nodes (\texttt{loc}) that are described as being in the \texttt{'Southeast'} and part of the \texttt{'United States'} (using either name or Wikidata description for robustness). This query exemplifies how ClimatePub4KG can help researchers find literature at the intersection of specific models, phenomena, and regional impacts.

\textbf{Persona 3: Enviromentalist}

Interest: Understanding the impact  of natural hazard on environment


\textbf{Natural Language Query 3:} ``Which papers mention the Pacific-North American (PNA) pattern in connection with locations in the United States?''

This query focuses on a specific teleconnection pattern (PNA) and its linkage to locations within the United States. The Cypher query is:


\begin{lstlisting}[language=Cypher, caption=Cypher Query 3]
MATCH (p:Paper)-[:Mention]->(t:Teleconnection 
    {Name:"PACIFIC_NORTH_AMERICAN_PNA_PATTERN"})
MATCH (t)-[:TargetsLocation]->(l:Location)
MATCH (p)-[:Mention]->(l)
WHERE l.wikidata_description CONTAINS "United States" OR 
      l.Name IN ["USA", "United States of America"]
RETURN p.title AS PaperTitle, t.Name AS TeleconnectionPattern, 
       l.Name AS Location
\end{lstlisting}

This query finds \texttt{Paper} nodes (\texttt{p}) mentioning the PNA pattern (\texttt{t}). It then identifies \texttt{Location} nodes (\texttt{l}) that are targeted by this pattern (i.e., the PNA influences these locations) and are also mentioned in the same paper. The query filters these locations to those whose Wikidata description contains ``United States'' or whose name explicitly indicates the USA. This showcases the ability to leverage external linked data (like Wikidata descriptions integrated into the ClimatePub4KG) and synonyms for more comprehensive geographical filtering.

These examples underscore how ClimatePub4KG, when queried with a language like Cypher, transforms information retrieval from a keyword-matching exercise into a sophisticated knowledge discovery process. Researchers can pose nuanced questions that reflect the complex interdependencies within climate science, receiving highly relevant and contextualized results. This capability enables more efficient literature reviews, facilitates the identification of connections between disparate pieces of information (e.g., linking specific models to observed phenomena in certain regions), and ultimately accelerates the pace of scientific research and understanding in this critical field. The ability to directly query the relationships *between* entities, rather than just the entities themselves, is a hallmark of KG-powered IR and a significant step towards more intelligent scientific information systems.

\section*{A comparison to ChatGPT-4o}

We pose natural language Query 1 to ChatGPT-4o and compare the resulting knowledge graph (KG) to the one generated by our system. As shown in Figures 1 and 2, the KG produced by ChatGPT lacks detailed information. It does not provide any citations (details not shown), and the direction of several relationships is incorrect. For example, the graph shows CMIP models influencing rainfall, rather than being evaluated against rainfall observations. While ChatGPT-4o identifies key concepts, our system captures richer and more accurate details—such as the specific version of the CMIP model (CMIP\_5), the papers that cite this version, and the region over which it was applied. These details are critical for scientific analysis and cannot be reliably replicated by ChatGPT, even when tested with two additional queries from earlier use cases.

\begin{figure}[ht]
  \centering
  \begin{minipage}[t]{0.48\linewidth}
    \centering
    \includegraphics[width=\linewidth]{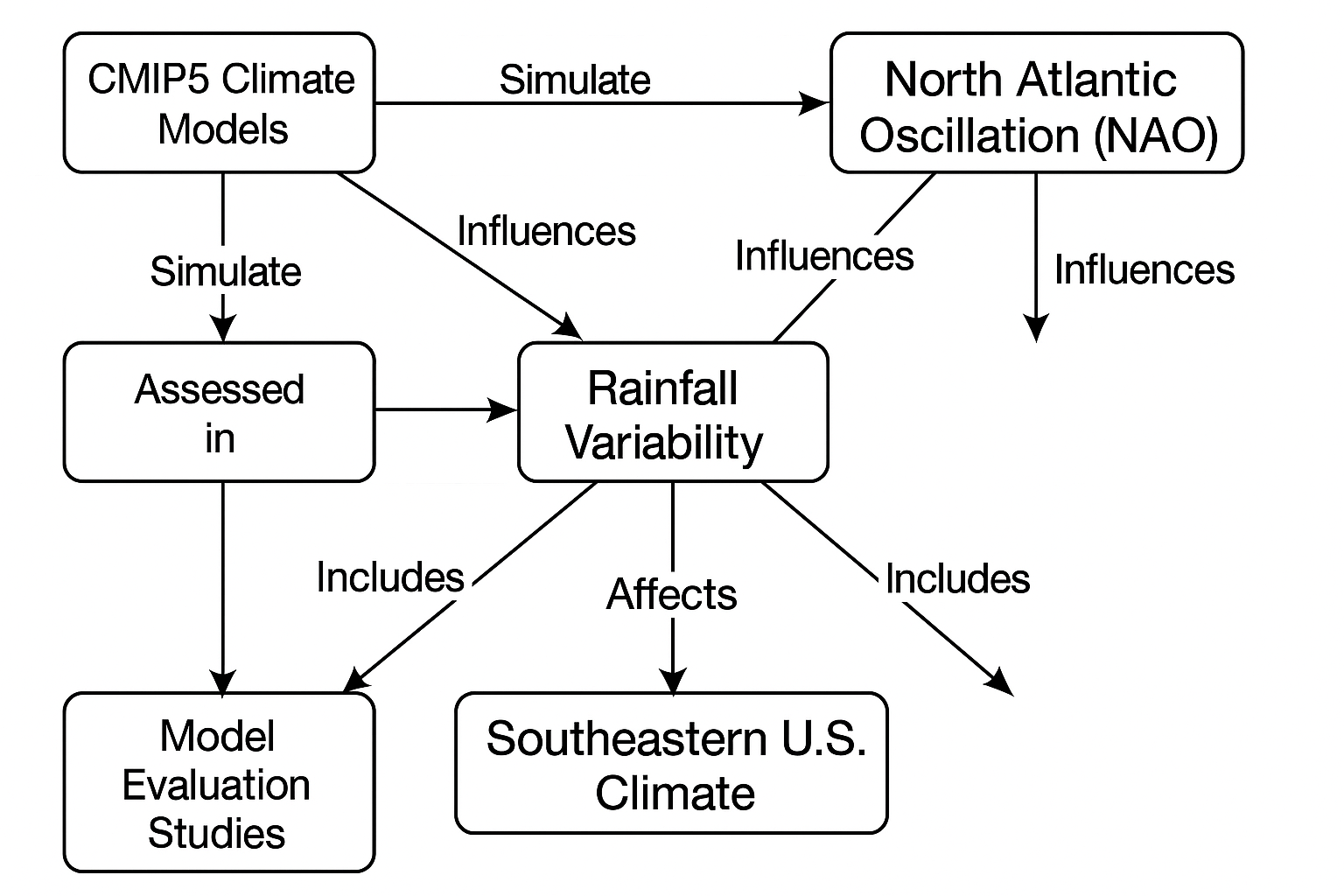}
    \caption{Graph generated by ChatGPT-4o for Query 1}
    \label{fig:chatgpt-kg}
  \end{minipage}
  \hfill
  \begin{minipage}[t]{0.48\linewidth}
    \centering
    \includegraphics[width=\linewidth]{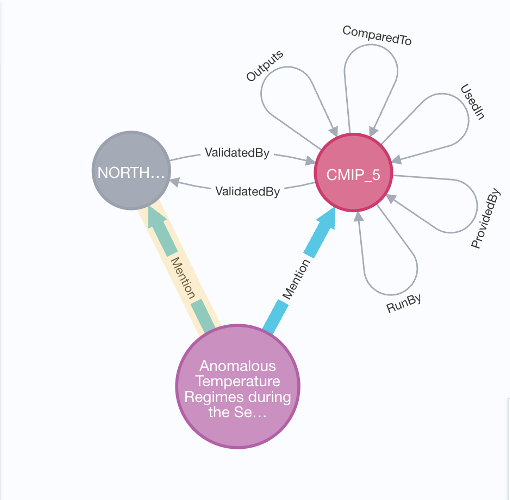}
    \caption{Graph generated by ClimatePub4KG for Query 1}
    \label{fig:climatepub4kg}
  \end{minipage}
\end{figure}

\section*{Applications and Implications for IR}
The integration of domain-specific Knowledge Graphs (KGs) into the information retrieval (IR) landscape, particularly within complex and data-rich scientific fields like climate science, carries profound implications for how researchers, policymakers, and the public discover, access, interpret, and utilize information. The use cases presented in the previous section, leveraging Cypher queries on ClimatePub4KG, offer a glimpse into a more powerful, precise, and context-aware paradigm for scientific inquiry. This KG-driven approach moves beyond traditional document retrieval, which often relies on keyword matching and statistical ranking, towards a system that understands and can navigate the intricate semantic relationships between scientific entities.

\textbf{Enhanced Precision and Recall in Scientific Search:} One of the most immediate applications is the significant enhancement of precision and recall in literature searches. By querying explicit relationships (e.g., a specific \texttt{Climate\_Model} \texttt{evaluatedWith} a particular \texttt{Observational\_Dataset} for a \texttt{Specific\_Region}), researchers can bypass the ambiguity of keyword searches and retrieve highly relevant documents or specific factual assertions. This is particularly crucial in climate science, where terminology can be highly nuanced and the same concept might be expressed in diverse ways. The ClimatePub4KG acts as a representation of knowledge, normalizing entities and relationships, thereby improving the consistency and comprehensiveness of search results.

\textbf{Facilitating Complex, Multi-hop Queries and Discovery of Non-Obvious Connections:} Traditional IR systems struggle with multi-hop queries that require traversing several relationships (e.g., ``Find all researchers who have published papers on \texttt{Topic\_A} using \texttt{Method\_X} that was originally developed for \texttt{Topic\_B}''). ClimatePub4KG excel at such queries. This capability allows for the discovery of non-obvious connections between different research areas, models, datasets, or phenomena. For instance, a researcher might uncover that a particular analytical method developed for atmospheric chemistry has been successfully applied to oceanographic data analysis, or that two seemingly unrelated climate phenomena share a common underlying driver, all through structured queries over the KG. This can spark new hypotheses and foster interdisciplinary collaboration.

\textbf{Complementing and Grounding LLMs:} 
While LLMs have shown remarkable capabilities in generating text and answering questions, they are prone to ``hallucinations''—generating plausible but factually incorrect information—especially in specialized scientific domains where training data might be sparse or highly technical. 
ClimatePub4KG can serve as a powerful grounding mechanism for LLMs. 
By retrieving factual information and contextual evidence from the ClimatePub4KG, LLMs can generate more accurate, reliable, and verifiable summaries, explanations, or answers to scientific queries.
Beyond factual grounding, knowledge graphs support structured reasoning. For example, Graph-Constrained Reasoning (GCR) integrates KG structure into LLM decoding using prefix-tree constraints, reducing hallucinations and enabling multi-hop reasoning \cite{ragReasoning}. 
Moreover, LLMs can be used to generate Cypher queries from natural language, enabling hybrid systems that retrieve precise answers from KGs before synthesis \cite{text2cypher}. 
ClimatePub4KG supports this functionality by allowing LLMs to convert user queries into Cypher, enabling structured retrieval and trustworthy, explainable answers in climate-related tasks.


\textbf{Supporting Systematic Reviews and Meta-Analyses:} The structured nature of ClimatePub4KG can significantly streamline the laborious process of conducting systematic reviews and meta-analyses. Researchers can programmatically query the ClimatePub4KG to identify all studies meeting specific criteria (e.g., using certain experimental designs, focusing on particular variables, or conducted within defined temporal and spatial boundaries), extract relevant data points, and analyze trends or inconsistencies across the literature. This can accelerate the synthesis of scientific evidence and the identification of knowledge gaps.



\textbf{Implications for IR Research:} The development and application of ClimatePub4KG like ours present several exciting research directions for the IR community. These include developing more effective natural language interfaces for querying ClimatePub4KG, creating robust evaluation benchmarks for KG-based retrieval in scientific domains, exploring techniques for automatically updating and expanding ClimatePub4KG as new research is published, and investigating methods for integrating KG-retrieved information with traditional document retrieval scores to provide richer, multi-faceted search results. Furthermore, research into KG embedding techniques can enable similarity-based searches and link prediction, further enhancing the discovery capabilities.


\section*{Conclusion and Future Work}

This paper has shown how a climate-specific Knowledge Graph (ClimatePub4KG), based on our prior work on ClimateIE and SciER, can serve as a practical tool for scientific information retrieval. Here we focused on the value of structured access to entities like models, datasets, and variables — and how that enables more precise, context-aware querying. The Cypher examples illustrate the types of questions this approach supports, from tracing model validation data to identifying research on specific phenomena.

Looking ahead, our focus is on improving usability and coverage. One direction is integrating the ClimatePub4KG with large language models to allow natural language queries, lowering the barrier for non-technical users. This is something that is already in progress and have advance in creating user interface for this purpose.  We also aim to extend the KG by continuously ingesting new literature and refining the taxonomy to reflect emerging concepts. Evaluation will be key: we are developing benchmark queries with domain experts to assess performance and usability.

Furthermore we plan to build user-friendly tools like search platforms, question-and-answer systems, and model comparison apps—that connect with ESGF, NASA Earthdata, and CMIP data to help scientists and decision-makers easily find, compare, and understand climate research.

We also see value in making the KG openly available, along with documentation and tools. Our aim is to support broader community use, encourage collaboration, and explore integration with ESGF, CMIP, and NASA Earthdata. This work is a step toward scalable, structured access to climate knowledge that helps researchers move faster and go deeper.

\section*{Acknowledgments}

This work was supported by the National Science Foundation awards III-2107213 and ITE-2333789. We also thank Aayush Acharya, Mykhailo Rudko and Quoc Truong Vinh Nguyen for their valuable contributions to our project. We also thank our reviewers for their feedback and comments.



\begin{thebibliography}{99} %

\bibitem{Xu2024} Xu, Zhentao, et al. "Retrieval-augmented generation with knowledge graphs for customer service question answering." Proceedings of the 47th International ACM SIGIR Conference on Research and Development in Information Retrieval. 2024.

\bibitem{graphRAG} Edge, Darren, et al. "From local to global: A graph rag approach to query-focused summarization." arXiv preprint arXiv:2404.16130 (2024).


\bibitem{Pan2024} Pan, G., Zhang, Y., Wang, S., et al. (2024). ClimateNER: Climate Science Information Extraction. *Prior Work Reference 1*. 

\bibitem{Zhang2024} Zhang, Y., Pan, G., Li, J., et al. (2024). SciER: An Entity and Relation Extraction Dataset for Datasets, Methods, and Tasks in Scientific Documents. *Prior Work Reference 2*. 

\bibitem{Lo2019S2ORC} Lo, K., Wang, L. L., Neumann, M., Kinney, R., and Weld, D. S. (2019). S2ORC: The Semantic Scholar Open Research Corpus. *Proceedings of the 2019 Conference on Empirical Methods in Natural Language Processing and the 9th International Joint Conference on Natural Language Processing (EMNLP-IJCNLP)*.

\bibitem{Sinha2015MAG} Sinha, A., Shen, Z., Song, Y., Ma, H., Eide, D., Hsu, B. P., and Wang, K. (2015). An Overview of Microsoft Academic Graph. *Proceedings of the 24th International Conference on World Wide Web (WWW ’15 Companion)*.

\bibitem{Tang2008AMiner} Tang, J., Zhang, J., Yao, L., Li, J., Zhang, L., and Su, Z. (2008). ArnetMiner: Extraction and Mining of Academic Social Networks. *Proceedings of the 14th ACM SIGKDD International Conference on Knowledge Discovery and Data Mining (KDD ’08)*.

\bibitem{Yu2018Spider} Yu, T., Zhang, R., Yang, K., Yasunaga, M., Wang, D., Li, Z., ... and Radev, D. (2018). Spider: A Large-Scale Human-Labeled Dataset for Complex and Cross-Domain Semantic Parsing and Text-to-SQL Task. *arXiv preprint arXiv:1809.08887*.

\bibitem{Hoffner2017KGQA} Höffner, K., Walter, S., Marx, E., Usbeck, R., Lehmann, J., \& Ngonga Ngomo, A. C. (2017). Survey on challenges of question answering in the web of data. *Semantic Web*, 8(6), 895-920.

\bibitem{Cui2020KGQA} Cui, W., Xiao, Y., Wang, H., Song, Y., Hwang, S. W., \& Wang, W. (2020). KBQA: Learning to Answer Questions over Knowledge Bases with Entity-Centric Graph Convolutions. *arXiv preprint arXiv:2002.00972*.


\bibitem{Ge2025Sustainability} Ge, R.; Xia, Y.; Ge, L.; Li, F. (2025). Knowledge Graph Analysis in Climate Action Research. *Sustainability*, 17, 371. https://doi.org/10.3390/su17010371

\bibitem{Islam2022AAAIClimateKB} Islam, M. S., Proma, A., Zhou, S., Hoque, E., \& Chen, L. (2022). ClimateKB: A Semantic KBS for Climate Change Related Knowledge Discovery. *Proceedings of the AAAI Conference on Artificial Intelligence*, 36(11), 12841-12843.

\bibitem{Wu2022ComputersGeosciences} Wu, J., Orlandi, F., O’Sullivan, D., and Devlin, K. (2022). A knowledge graph based approach for climate change literature review. *Computers and Geosciences*, 167, 105198. https://doi.org/10.1016/j.cageo.2022.105198

\bibitem{Tu2024ICCSA} Tu, S., Zhuang, W., \& Ren, F. (2024). Construction of Knowledge Graph for Climate Change and Its Application in Q\&A System. In: Gervasi, O., Murgante, B., Misra, S., Rocha, A.M.A.C., Garau, C. (eds) Computational Science and Its Applications – ICCSA 2024. ICCSA 2024. Lecture Notes in Computer Science, vol 14619. Springer, Cham. https://doi.org/10.1007/978-981-96-3433-0\_22

\bibitem{Zhu2025Arxiv} Zhu, R., Shimizu, C., Stephen, J., et al. (2025). ClimateKG: A Climate Science Knowledge Graph. *arXiv preprint arXiv:2502.13874v1*.

\bibitem{Zhao2024SemanticScholar} Zhao, Y., Guo, J., Bao, C., Li, Y., \& Wang, B. (2024). Climate Change Knowledge Graph Construction and Application: A Survey. *Semantic Scholar*. (Example, actual citation details may vary for preprints/surveys on Semantic Scholar without formal publication venues yet).

\bibitem{Wu2024IEEEAccess} Wu, J., Orlandi, F., O’Sullivan, D., \& Devlin, K. (2024). A Knowledge Graph-Based Approach for Climate Change Literature Review. *IEEE Access*, 12, 10719340. (Note: This seems to be a duplicate or very similar to the Computers \& Geosciences one, common in pre-print to journal pipelines. Using the IEEE Access one as an example of a different venue if it were distinct).

\bibitem{Fotopoulou2022Frontiers} Fotopoulou, E., Mandilara, I., Tsalis, N., et al. (2022). Knowledge graphs for climate change and health nexus: A systematic review. *Frontiers in Environmental Science*, 10, 1003599. https://doi.org/10.3389/fenvs.2022.1003599

\bibitem{Peng2023AIR} Peng, C., Xia, F., Naseriparsa, M., \& Osborne, F. (2023). Knowledge Graphs: Opportunities and Challenges. *Artificial Intelligence Review*, 56, 13071–13102. https://doi.org/10.1007/s10462-023-10465-9

\bibitem{Kejriwal2022Information} Kejriwal, M. (2022). Knowledge Graphs: A Practical Review of the Research Landscape. *Information*, 13(4), 161. https://doi.org/10.3390/info13040161

\bibitem{AbuSalih2024Heliyon} Abu-Salih, B., \& Alotaibi, S. (2024). Knowledge graph for enhancing scientific research: A systematic literature review. *Heliyon*, 10(8), e29034. https://doi.org/10.1016/j.heliyon.2024.e29034

\bibitem{Callahan2024NatureSciData} Callahan, T. J., Tripodi, I. J., et al. (2024). The Biomedical Data Translator knowledge graph. *Scientific Data*, 11, 317. https://doi.org/10.1038/s41597-024-03171-w

\bibitem{Manghi2024QSS} Manghi, P., Atzori, C., Bardi, A., et al. (2024). Challenges in building scholarly knowledge graphs. *Quantitative Science Studies*, 5(4), 991-1020. https://doi.org/10.1162/qss\_a\_00328

\bibitem{dAquin2025WebSem} d’Aquin, M. (2025). On the role of knowledge graphs in AI-based scientific discovery. *Journal of Web Semantics*, 84, 100854. https://doi.org/10.1016/j.websem.2024.100854

\bibitem{Xu2022Neurocomputing} Xu, M., Du, J., Xue, Z., Kou, G., \& Zhang, C. (2022). Semantic information retrieval for scientific literature: A review. *Neurocomputing*, 492, 336-351. https://doi.org/10.1016/j.neucom.2021.11.063

\bibitem{Sharma2023Arxiv} Sharma, S., \& Jain, S. (2023). Comprehensive Review on Semantic Information Retrieval and Ontology Engineering. *arXiv preprint arXiv:2307.13427*.

\bibitem{DynClean} Zhang, Q., Pan, H., Chen, Z., Latecki, L. J., Caragea, C., \& Dragut, E. (2025, April). DynClean: Training Dynamics-based Label Cleaning for Distantly-Supervised Named Entity Recognition. In Findings of the Association for Computational Linguistics: NAACL 2025 (pp. 2540-2556).

\bibitem{Zhu2024ArxivKGCReasoning} Zhu, Y., Wang, X., Chen, J., Qin, Y., \& Liu, K. (2024). A Survey on Knowledge Graph Construction and Reasoning. *arXiv preprint arXiv:2305.13168v2*. (Note: Year might be 2023 for original submission, using 2024 for v2 as an example if it was updated then).

\bibitem{Sahlab2022Arxiv} Sahlab, N., Kahoul, H., Jazdi, N., \& Weyrich, M. (2022). A Knowledge Graph-Based Method for Automating Systematic Literature Reviews. *arXiv preprint arXiv:2208.02334*.

\bibitem{Liang2023ArxivKGReasoningSurvey} Liang, K., Meng, L., Liu, M., et al. (2023). A Survey on Knowledge Graph Reasoning. *arXiv preprint arXiv:2212.05767*.

\bibitem{pan2024climatener}
Pan, H., Zhang, Q., Adamu, M., Dragut, E., \& Latecki, L. J. (2024). ClimateNER: Climate Science Information Extraction. *Proceedings of the 2024 Conference on Empirical Methods in Natural Language Processing (EMNLP)*.

\bibitem{zhang2024scier}
Zhang, Q., Chen, Z., Pan, H., Caragea, C., Latecki, L. J., \& Dragut, E. (2024). SciER: An Entity and Relation Extraction Dataset for Datasets, Methods, and Tasks in Scientific Documents. *Proceedings of the 2024 Conference on Empirical Methods in Natural Language Processing (EMNLP)*, 13083–13100.

\bibitem{wu2022linkclimate}
Wu, J., Orlandi, F., O'Sullivan, D., \& Dev, S. (2022). LinkClimate: An Interoperable Knowledge Graph Platform for Climate Data. *Environmental Modelling \& Software*, 152, 105384.

\bibitem{reinanda2020knowledge}
Reinanda, R., Meij, E., \& de Rijke, M. (2020). Knowledge Graphs: An Information Retrieval Perspective. *arXiv preprint arXiv:2003.02320*.

\bibitem{defila2011defining}
Defila, R., \& Di Giulio, A. (2011). Defining Terms for Integrated (Multi-Inter-Trans-Disciplinary) Sustainability Research. *Sustainability*, 3(8), 1090–1113.

\bibitem{metoffice2022interdisciplinary}
Met Office. (2022). An Interdisciplinary Approach for Climate Risk Analysis and Communication. Retrieved from \url{https://www.metoffice.gov.uk/}.

\bibitem{schmidt2018reproducibility}
Schmidt, G. (2018). Reproducibility and Replication in Climate Science. Retrieved from \url{https://www.realclimate.org/}.

\bibitem{rupprecht2020ursprung}
Rupprecht, D., et al. (2020). Improving Reproducibility of Data Science Pipelines through Transparent Provenance Collection. *Proceedings of the VLDB Endowment*, 13(12), 3354–3368.

\bibitem{Sahlab2022Arxiv}
Sahlab, N., Kahoul, H., Jazdi, N., \& Weyrich, M. (2022). A Knowledge Graph-Based Method for Automating Systematic Literature Reviews. *arXiv preprint arXiv:2208.02334*.

\bibitem{Liang2023ArxivKGReasoningSurvey}
Liang, K., Meng, L., Liu, M., et al. (2023). A Survey on Knowledge Graph Reasoning. *arXiv preprint arXiv:2212.05767*.

\bibitem{MoreRef1} Author, A. A. (Year). Title of More Reference 1. *Journal of More References*, Vol(Iss), Pages.
\bibitem{MoreRef2} Author, B. B. (Year). Title of More Reference 2. *Conference on More References*.
\bibitem{MoreRef3} Author, C. C. (Year). Title of More Reference 3. *Book Chapter in More References Book*.
\bibitem{MoreRef4} Author, D. D. (Year). Title of More Reference 4. *Another Journal*, Vol(Iss), Pages.
\bibitem{MoreRef5} Author, E. E. (Year). Title of More Reference 5. *Yet Another Conference*.
\bibitem{Kuma2023JAMES}
Kuma, P., Bender, F.~A.-M., \& Jönsson, A.~R. (2023). Climate model code genealogy and its relation to climate feedbacks and sensitivity. \textit{Journal of Advances in Modeling Earth Systems}, 15, e2022MS003588. https://doi.org/10.1029/2022MS003588

\bibitem{lo2019s2orc}
Lo, K., Wang, L. L., Neumann, M., Kinney, R., \& Weld, D. S. (2019). S2ORC: The Semantic Scholar Open Research Corpus. \textit{Proceedings of the 2019 Conference on Empirical Methods in Natural Language Processing}.

\bibitem{prieme2022openalex}
Priem, J., \& Piwowar, H. (2022). OpenAlex: A fully-open catalog of the global research system. Retrieved from \url{https://docs.openalex.org/}

\bibitem{tang2008arnetminer}
Tang, J., Zhang, J., Yao, L., Li, J., Zhang, L., \& Su, Z. (2008). ArnetMiner: Extraction and Mining of Academic Social Networks. \textit{Proceedings of the 14th ACM SIGKDD International Conference on Knowledge Discovery and Data Mining}, 990–998.

\bibitem{MoreRef6} Author, F. F. (Year). Title of More Reference 6. *Final Reference Journal*.

\bibitem{ragReasoning} Luo, Linhao, et al. "Graph-constrained reasoning: Faithful reasoning on knowledge graphs with large language models." arXiv preprint arXiv:2410.13080 (2024).

\bibitem{text2cypher} Ozsoy, Makbule Gulcin, et al. "Text2Cypher: Bridging Natural Language and Graph Databases." arXiv preprint arXiv:2412.10064 (2024).

\bibitem{dragut2024climatepub4kg}
Dragut, E., Latecki, L., Rudko, M., Adamu, M., Zhang, Q., \& Pan, J. (2024). ClimatePub4KG: Toward a Knowledge Graph to Support Evaluation and Development of Climate Models. *AGU Fall Meeting Abstracts*, IN11A-01.

\bibitem{kruger2006nexrad}
Kruger, A., Lawrence, R., \& Dragut, E. C. (2006). Building a terabyte NEXRAD radar database for hydrometeorology research. *Computers \& Geosciences*, 32(2), 247–258.

\bibitem{pan2025taxonomykg}
Pan, H., Zhang, Q., Adamu, M., Dragut, E. C., \& Latecki, L. J. (2025). Taxonomy-Driven Knowledge Graph Construction for Domain-Specific Scientific Applications. *Findings of the Association for Computational Linguistics: ACL 2025*.

\bibitem{pan2025climateie}
Pan, H., Adamu, M., Zhang, Q., Dragut, E. C., \& Latecki, L. J. (2025). ClimateIE: A Dataset for Climate Science Information Extraction. *Proceedings of the ClimateNLP Workshop at ACL 2025*.


\end{thebibliography}
\end{document}